\documentclass[letterpaper, 10 pt, journal, twoside]{IEEEtran}

\usepackage{graphics} 
\usepackage{subfigure}
\usepackage{epsfig} 
\usepackage{amsmath} 
\usepackage{amssymb}  
\usepackage{booktabs}
\usepackage{algorithm}
\usepackage{algpseudocode}
\usepackage{cite}
\usepackage{url}
\usepackage{xcolor}

\begin{document}

\title{From Transcripts to AI Agents: Knowledge Extraction, RAG Integration, and Robust Evaluation of Conversational AI Assistants}

\author{Krittin Pachtrachai, Petmongkon Pornpichitsuwan, Wachiravit Modecrua and Touchapon Kraisingkorn
\thanks{Krittin Pachtrachai, Petmongkon Pornpichitsuwan, Wachiravit Modecrua and Touchapon Kraisingkorn are with Amity Research and Application Center (ARAC), Amity Solutions, Bangkok, Thailand.
        {\tt\footnotesize \{krittin.pac; petmongkon; wachiravit\}@amitysolutions.com,
touchapon@amity.co}}
}

\maketitle

\bstctlcite{BSTcontrol}

\begin{abstract}
Building reliable conversational AI assistants for customer-facing industries remains challenging due to noisy conversational data, fragmented knowledge, and the requirement for accurate human hand-off—particularly in domains that depend heavily on real-time information. This paper presents an end-to-end framework for constructing and evaluating a conversational AI assistant directly from historical call transcripts. Incoming transcripts are first graded using a simplified adaptation of the PIPA framework, focusing on observation alignment and appropriate response behavior, and are filtered to retain only high-quality interactions exhibiting coherent flow and effective human agent responses. Structured knowledge is then extracted from curated transcripts using large language models (LLMs) and deployed as the sole grounding source in a Retrieval-Augmented Generation (RAG) pipeline. Assistant behavior is governed through systematic prompt tuning, progressing from monolithic prompts to lean, modular, and governed designs that ensure consistency, safety, and controllable execution. Evaluation is conducted using a transcript-grounded user simulator, enabling quantitative measurement of call coverage, factual accuracy, and human escalation behavior. Additional red teaming assesses robustness against prompt injection, out-of-scope, and out-of-context attacks. Experiments are conducted in the Real Estate and Specialist Recruitment domains, which are intentionally challenging and currently suboptimal for automation due to their reliance on real-time data. Despite these constraints, the assistant autonomously handles approximately 30\% of calls, achieves near-perfect factual accuracy and rejection behavior, and demonstrates strong robustness under adversarial testing.
\end{abstract}


\IEEEpeerreviewmaketitle

\section{Introduction}
\label{sec:intro}

Large Language Models (LLMs) have rapidly transitioned from research artifacts to widely deployed AI assistants embedded in everyday digital experiences. Advances in transformer architectures and large-scale pretraining have enabled LLMs to perform natural language understanding and generation at human-like levels, supporting tasks such as information retrieval, writing assistance, code generation, and conversational interaction \cite{vaswani2017attention,devlin2019bert,brown2020language}. Through instruction tuning and reinforcement learning from human feedback, modern LLMs have further improved alignment and usability, leading to their integration into search engines, productivity tools, and conversational interfaces encountered in daily life \cite{bommasani2021foundation,wei2022chain,ouyang2022training}. This transition has also sparked renewed interest in the design and classification of AI agents and agentic systems, which explore modular task automation and multi-agent coordination beyond single-turn interactions \cite{Sapkota_2026,yang2025surveyaiagentprotocols}.

One of the most impactful application domains for LLM-based assistants is customer service and call center operations, where a substantial proportion of interactions consist of repetitive, information-seeking queries. Prior studies show that conversational AI can reduce handling time, improve consistency, and significantly lower operational costs when deployed appropriately \cite{huang2018service,davenport2018artificial, kaewtawee2025cloningconversationalvoiceai}. Economic analysis further suggests that automation technologies can alleviate labor shortages and scale service capacity, creating strong incentives for adoption in high-volume service environments such as call centers \cite{autor2015work,frey2017future,acemoglu2020robots}.

However, the adoption of AI assistants also introduces human and organizational challenges. Workers may experience anxiety related to job displacement, increased monitoring, or intensified performance expectations, leading to stress and resistance toward AI-enabled workflows \cite{frey2017future,acemoglu2020robots}. Research in human–automation interaction highlights risks such as deskilling, loss of situational awareness, and increased cognitive load when automation is poorly designed or over-trusted \cite{bainbridge1983ironies,parasuraman1997humans,seeber2020machines}.

In contrast, a growing body of evidence demonstrates that AI-assisted work—where AI augments rather than replaces human labor—can yield substantial productivity and quality gains. Human–AI collaboration has been shown to outperform either humans or AI alone across a range of tasks, including decision-making and service delivery \cite{brynjolfsson2018augmentation,kamar2016hybrid}. In call center contexts, AI assistance can support real-time knowledge retrieval, response suggestion, and escalation decisions, enabling agents to focus on complex and empathy-driven interactions \cite{amershi2019guidelines,dellermann2019hybrid,seeber2020machines}.

Motivated by this paradigm, we introduce an AI assistant designed to provide general domain knowledge derived from real conversational transcripts, with the explicit goal of reducing human workload rather than fully automating service interactions. By handling routine informational requests and accurately transferring unresolved cases to human agents, the system establishes a foundation for scalable and sustainable human–AI collaboration \cite{autor2015work,davenport2018humanmachine,shneiderman2020human}.

This paper presents a transcript-driven framework for building and evaluating such an assistant. Our approach extracts structured knowledge from curated conversations and deploys it via a Retrieval-Augmented Generation (RAG) pipeline, enabling human-like information delivery grounded in real operational data. We further introduce a realistic evaluation methodology based on user simulation and red teaming to assess coverage, accuracy, and robustness, demonstrating a practical pathway for deploying reliable AI assistants in challenging real-world domains \cite{lewis2020retrieval,karpukhin2020dense,izacard2022fewshot,gao2023survey}.
\section{Related Works}
\label{sec:related}

\subsection{Research on Voice-Based Conversational AI}
Academic research on conversational AI has historically focused on text-based dialogue systems, where language understanding, dialogue state tracking, and response generation could be studied in isolation. With advances in deep learning and large-scale speech datasets, recent work has moved toward end-to-end spoken dialogue systems integrating automatic speech recognition (ASR), natural language understanding, dialogue management, and text-to-speech (TTS).

One prominent example is Google Duplex, which demonstrated that neural conversational systems can conduct natural-sounding telephone conversations, including turn-taking, hesitations, and contextual grounding \cite{levine2020duplex}. While Duplex showcased the feasibility of highly natural voice interactions, it relied on tightly constrained task domains (e.g., restaurant reservations) and significant manual engineering, limiting its generalizability.

Hybrid human–AI systems have also been explored to address robustness and deployment challenges. Evorus proposed a crowd-powered conversational assistant that gradually automates responses by learning from human interventions \cite{huang2018evorus}. This approach improves early-stage reliability and enables data-driven automation but introduces latency and operational cost, making it less suitable for real-time, large-scale voice interactions.

Parallel to dialogue system research, RAG has become a dominant paradigm for grounding large language models in external knowledge sources. Techniques such as Dense Passage Retrieval and retrieval-augmented sequence-to-sequence generation significantly improve factual accuracy in knowledge-intensive tasks \cite{karpukhin2020dense,lewis2020retrieval}. However, most RAG research assumes clean, well-structured knowledge bases and text-based interaction, leaving open challenges when applied to noisy, real-world voice transcripts and dynamic conversational contexts.

\subsection{Commercial Voice AI Platforms}

Commercial voice AI platforms have operationalized conversational AI for customer service and call center automation, prioritizing reliability, latency, and enterprise integration over experimental flexibility.

PolyAI focuses on fully automated, enterprise-grade voice assistants capable of handling complex, multi-turn conversations in customer support settings. Its strengths include robust dialogue flow management, multilingual capabilities, and strong performance in intent-heavy scenarios such as banking or hospitality \cite{polyai2023whitepaper}. However, PolyAI systems typically require extensive domain-specific configuration and curated training data, making rapid cross-domain deployment costly and time-consuming.

Kore.ai provides a comprehensive conversational AI platform supporting both voice and text channels, with strong emphasis on agent assist, analytics, and workflow orchestration \cite{koreai2022platform}. Its modular architecture allows enterprises to integrate conversational AI into existing systems with relative ease. The main limitation is that Kore.ai often functions as a co-pilot rather than a fully autonomous voice agent, relying on predefined flows and human oversight for complex or ambiguous interactions.

Emerging platforms such as Retell AI, Sierra, and Decagon leverage large language models to enable more flexible, natural voice interactions with minimal scripting. These systems excel in low-latency response generation and rapid prototyping, enabling near-human conversational experiences \cite{retell2023docs,sierra2023overview,decagon2023system}. Nevertheless, their reliance on prompt-based reasoning introduces challenges related to hallucination, consistency, and out-of-scope query handling. Ensuring safe escalation to human agents and maintaining deterministic behavior remain open operational concerns.

Across commercial systems, a common limitation is dependence on manually curated knowledge sources, rigid business logic, or real-time backend integrations. While effective in stable domains, these constraints reduce adaptability when conversational patterns evolve or when historical call data is underutilized.

Both academic and commercial approaches have made significant progress in voice-based conversational AI. Research systems emphasize innovation and theoretical grounding but often lack scalability and real-world robustness. Commercial platforms achieve production readiness but depend heavily on domain engineering and predefined logic. These gaps motivate approaches that automatically extract structured knowledge from historical call transcripts, integrate retrieval-augmented reasoning, and evaluate performance under realistic conversational conditions.
\section{System Overview}
\label{sec:methods}

\begin{figure}[t]
\centering
\includegraphics[width=\linewidth]{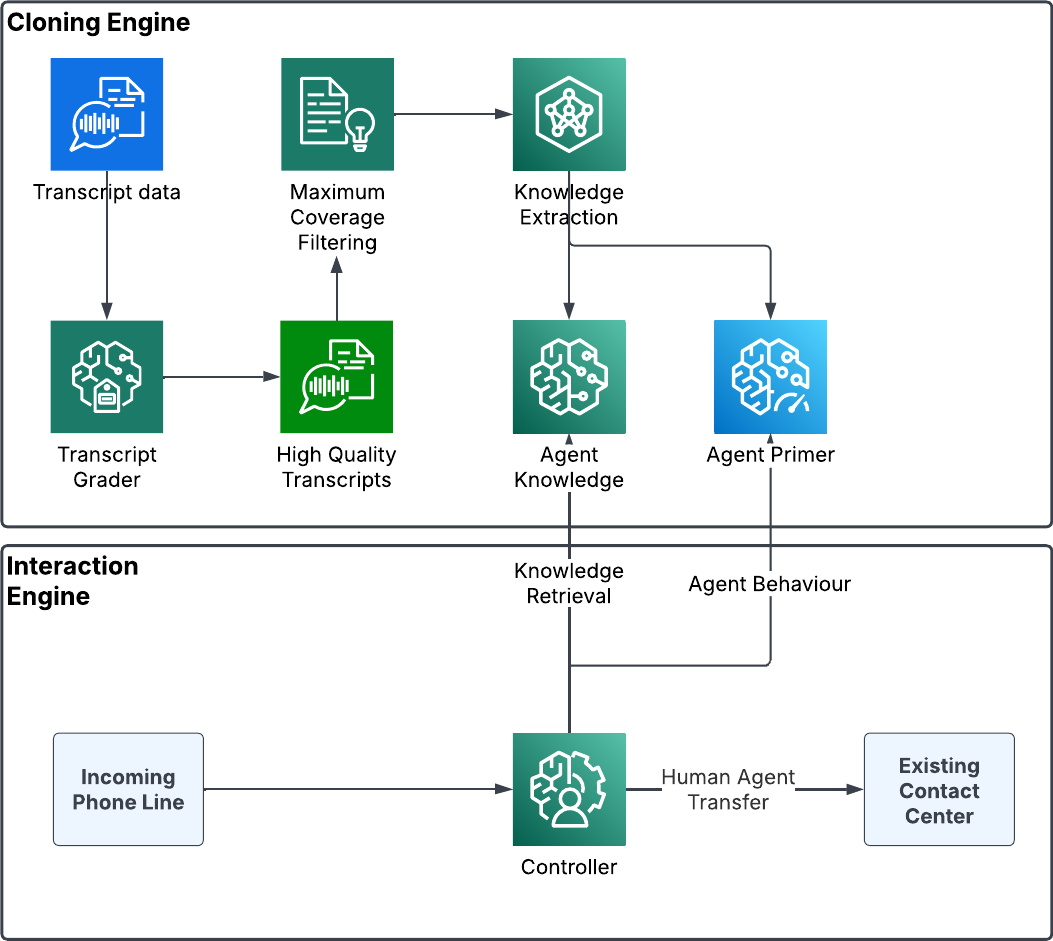}
\caption{Overview of the proposed transcript-to-agent pipeline. The framework illustrates transcript grading and filtering, strategic selection for knowledge coverage, knowledge extraction, prompt tuning, RAG-based response generation, and human agent transfer when escalation is required.}
\label{fig:system_overview}
\end{figure}

This section presents an overview of the proposed framework for constructing a conversational AI assistant from historical call transcripts. As illustrated in Figure \ref{fig:system_overview}, the system is designed to transform unstructured human–human conversations into a deployable AI agent through a structured pipeline. The framework integrates transcript grading and filtering for quality control, strategic selection of conversations to maximize knowledge coverage, systematic knowledge extraction, and behavior conditioning via prompt engineering. Each component is modular and designed to support scalable development and rigorous evaluation of conversational AI assistants.

\subsection{Grading}

Real-world call transcripts exhibit substantial variability in quality, structure, and informational reliability. Factors such as incomplete conversations, contradictory statements, non-interactive system messages, and varying levels of human agent expertise necessitate an explicit grading and filtering stage prior to knowledge extraction. To address this, we adopt a transcript evaluation framework inspired by recent work on conversational policy analysis and observation alignment \cite{kim2025pipa}, while tailoring it to the requirements of large-scale, offline transcript processing.

Specifically, our framework incorporates \emph{Observation Alignment} as a core criterion, which assesses whether assistant responses are grounded in observable conversational evidence rather than unsupported assumptions or hallucinated content. Let \( a_i \) denote the \(i\)-th assistant utterance in a transcript and \( N \) be the total number of assistant utterances evaluated. The observation alignment score \( A_{\text{obs}} \) is defined as
\begin{equation}
A_{\text{obs}} = \frac{1}{N} \sum_{i=1}^{N} \mathbb{I}\left( a_i \;\text{is aligned with observable evidence} \right),
\end{equation}
where \( \mathbb{I}(\cdot) \) is an indicator function that evaluates to 1 if the assistant response is consistent with user-provided information and prior observable context, and 0 otherwise.

In addition to Observation Alignment, we introduce a novel behavioral criterion termed \emph{Appropriate Response (P)}. This criterion evaluates assistant behavior at the utterance level and is applied exclusively to assistant turns inferred from the transcript. Each assistant message is judged independently for politeness, respectful and professional tone, clarity, and self-contained completeness, without assessing factual correctness or long-term state consistency. The Appropriate Response score \( A_{\text{P}} \) is defined as
\begin{equation}
A_{\text{P}} = \frac{1}{N} \sum_{i=1}^{N} \mathbb{I}\left( a_i \;\text{meets Appropriate Response (P)} \right),
\end{equation}
where the indicator function returns 1 if the utterance satisfies the behavioral and communicative criteria of P, and 0 otherwise. Appropriate Response (P) evaluates the intrinsic quality of assistant utterances independently of factual correctness or dialogue context. A response is considered appropriate if it is polite, professional in tone, and expressed as a clear, self-contained statement that can be understood in isolation. The criterion emphasizes completeness and coherence, allowing explicit acknowledgments of uncertainty or limitations when they are communicated clearly and respectfully. Responses are marked inappropriate when they are fragmentary, ambiguous, internally inconsistent, unintelligible, or consist solely of non-substantive pleasantries or system-level artifacts. By operating at the utterance level and excluding user turns, this criterion measures the assistant’s ability to produce consistently well-formed and deployable responses suitable for customer-facing conversational systems.

Assistant turns are identified through role inference based on conversational cues, while user utterances, system announcements, and low-confidence or unreadable segments are excluded from evaluation. Each eligible assistant utterance is assigned a binary alignment decision under both criteria. The final transcript score is computed by jointly considering \( A_{\text{obs}} \) and \( A_{\text{P}} \), and only transcripts exceeding predefined thresholds for both scores are retained for downstream processing.

It is important to note that both Observation Alignment and Appropriate Response are intentionally defined as simplified, interpretable criteria to enable scalable evaluation across large transcript corpora. These criteria represent a foundational subset of a more comprehensive behavioral and reasoning assessment framework, which may incorporate richer policy dimensions, graded scoring, and temporal dependencies in future work.

Finally, the evaluation under both Observation Alignment and Appropriate Response (P) is conducted using a \emph{large language model acting as an automatic evaluator} (LLM-as-a-judge). Recent studies have demonstrated that LLM-based evaluators achieve strong agreement with human judgments for qualitative, behavioral, and instruction-following assessments, while offering substantially improved scalability and consistency compared to manual annotation \cite{zheng2023judging,liu2023geval,dubois2023alpacaeval,wang2024judgelm}. Prior work further shows that LLM-as-a-judge frameworks correlate well with expert evaluations across diverse conversational tasks and domains, making them suitable for large-scale offline evaluation where human labeling is impractical. In this work, the LLM judge is used exclusively for binary decision-making under clearly defined criteria, which further mitigates evaluator bias and enhances reproducibility.

\subsection{Filtering and Knowledge Coverage Maximization}

Following transcript grading, a second filtering stage is applied to select a subset of conversations that maximizes knowledge utility for downstream extraction. First, only transcripts whose normalized grading score exceeds a predefined threshold of 0.8 are retained. This threshold is empirically chosen to balance transcript quality and data availability, ensuring that retained conversations exhibit both strong observation alignment and appropriate assistant behavior.

The choice of the 0.8 threshold is motivated by practical characteristics of real-world call data. The transcripts are produced via automatic diarization of call recordings, which is inherently imperfect and often introduces artifacts such as role misattribution, speaker switching, incomplete utterances, repetitive enunciations, and incoherent turn boundaries caused by background noise or unstable phone signals. Lower-scoring transcripts frequently contain contradictory statements, fragmented exchanges, or non-interactive system segments that degrade the reliability of extracted knowledge. In contrast, transcripts exceeding the 0.8 threshold consistently demonstrate coherent conversational flow, stable speaker roles, non-conflicting information, and sufficiently rich domain content.

By enforcing this threshold, the filtering stage ensures that subsequent knowledge extraction operates on conversations that are both structurally reliable and semantically informative. Although this constraint reduces the total number of usable transcripts, it significantly improves the signal-to-noise ratio of the resulting knowledge base, thereby enhancing the quality, consistency, and usability of the extracted knowledge for retrieval-augmented generation in later stages of the pipeline.

Each retained transcript is then annotated with a set of knowledge topics reflecting the issues, procedures, or domain concepts discussed in the conversation. Topic assignment is performed using a large language model, which analyzes the transcript holistically and identifies salient knowledge categories present in the dialogue. These topics are treated as abstract representations of knowledge units, allowing heterogeneous conversational content to be compared and aggregated across transcripts.

Given the resulting transcript–topic associations, the selection of conversations for knowledge extraction is formulated as a \emph{maximum coverage} problem. Let \( \mathcal{T} = \{t_1, \dots, t_M\} \) denote the set of all identified knowledge topics, and let each transcript \( c_i \) cover a subset \( \mathcal{T}_i \subseteq \mathcal{T} \). The objective is to select a subset of transcripts \( \mathcal{C}^\ast \) that maximizes the union of covered topics,
\begin{equation}
\mathcal{C}^\ast = \arg\max_{\mathcal{C}' \subseteq \mathcal{C}} \left| \bigcup_{c_i \in \mathcal{C}'} \mathcal{T}_i \right|.
\end{equation}

To efficiently approximate this objective, we apply a greedy selection algorithm that iteratively adds the transcript contributing the largest number of previously uncovered topics. While the maximum coverage problem is NP-hard, the greedy strategy provides a well-known approximation guarantee and performs effectively in practice. This approach prioritizes diversity and breadth of knowledge rather than redundancy, ensuring that the selected transcripts collectively capture a wide range of domain concepts.

By combining quality-based thresholding with coverage-aware selection, this filtering stage produces a compact yet information-rich set of transcripts. This design maximizes the amount of extractable knowledge in the subsequent knowledge extraction and structuring stage, while controlling data volume and minimizing redundant conversational patterns.

\subsection{Knowledge Extraction and Representation}

Following transcript filtering, the selected conversations are processed to extract structured domain knowledge. For this step, we adopt the knowledge extraction methodology proposed in \cite{kaewtawee2025cloningconversationalvoiceai}, which demonstrates how large language models can distill procedural and factual knowledge from conversational data. In the original formulation, the extracted knowledge is directly incorporated into a behavioral primer that conditions the assistant’s responses.

In our system, however, the extracted knowledge is used exclusively as an external knowledge source rather than being embedded into the system prompt. This design choice is motivated by practical limitations of long system prompts in large language models. Prior work has shown that excessively long prompts can degrade model performance, particularly when relevant information is diluted by irrelevant or distant context, a phenomenon commonly referred to as the ``lost-in-the-middle'' effect \cite{liu2024lost, du-etal-2025-context, press2022train}. As a result, encoding large volumes of domain knowledge directly into the prompt may reduce response accuracy and consistency, especially in multi-turn conversational settings.

Accordingly, we retain only concise behavioral instructions in the system prompt and store the extracted knowledge in a structured repository suitable for retrieval-augmented generation (RAG). During inference, relevant knowledge snippets are dynamically retrieved based on the user query and injected into the model context as needed. This approach decouples behavioral conditioning from domain knowledge grounding, allowing the assistant to access a broad and evolving knowledge base without incurring the drawbacks associated with long, static prompts.

By leveraging transcript-derived knowledge as a retrieval source rather than prompt content, the proposed system preserves prompt efficiency while maintaining factual grounding. This design also enables scalable updates to the knowledge base as new transcripts become available, without requiring prompt re-engineering.

\subsection{Prompt Tuning and Agent Behavior Design}

Prompt tuning is used to define the high-level behavior of the conversational agent independently of domain-specific knowledge. Rather than embedding task-specific rules or factual content into the prompt, we focus on optimizing a \emph{generic behavioral prompt} that can be applied across multiple use cases and domains. This design enables the agent to operate consistently while relying on external knowledge retrieval for factual grounding.

We distinguish between two broad classes of conversational agents based on interaction style: \emph{active agents} and \emph{reactive agents}. Active agents proactively drive conversations toward predefined objectives, such as sales outreach or feedback collection, whereas reactive agents primarily respond to user-initiated requests, including customer service, technical assistance, and receptionist-style interactions. In this work, we focus exclusively on prompt design for reactive agents, as they represent the most common and operationally critical use case in call center environments.

The prompt is constructed to remain concise, and non-redundant, avoiding excessive length that could dilute relevant instructions or degrade model performance. Prior studies have shown that long prompts can impair information utilization and increase error rates due to positional bias and context saturation \cite{liu2024lost,du-etal-2025-context}. To further ensure reliable behavior, the prompt is designed to be internally consistent and free of conflicting instructions, as contradictory directives have been shown to reduce instruction-following accuracy in large language models \cite{ouyang2022training}.

During development, we iteratively explored several prompt architectures, reflecting increasing levels of modularity, abstraction, and governance. These variants were not all intended as final designs but served to explore trade-offs between controllability, generality, and operational robustness.

\subsubsection{Early Escalation}
The initial design prioritized human escalation, embedding detailed domain knowledge directly within the system prompt. While conservative and safe, this approach resulted in long prompts and limited scalability across domains.

\subsubsection{Low-autonomy RAG}
Domain knowledge was removed from the prompt and delegated to a RAG pipeline, while escalation behavior remained conservative. This reduced prompt length but still constrained agent autonomy. This is to prove that our development conforms to the literature \cite{liu2024lost, bommasani2021foundation}.

\subsubsection{Modular Orchestration}
The system prompt was streamlined such that the number of lines is reduced significantly. This is done by separating invariant behavioral logic from domain-specific components. Approximately 90\% of the prompt became reusable across industries and roles, with behavior defined through explicit step-by-step action sequences.

\subsubsection{Programmatic Orchestration}
Building on modular orchestration, action steps were expressed in Python-like pseudo-code to reduce ambiguity and encourage deterministic execution patterns.

\subsubsection{YAML-based Control}
Prompt sections were further formalized using YAML-style structures, improving readability, consistency, and machine interpretability.

\subsubsection{Protocol-based Orchestration}
The prompt was decomposed into minimal, reusable behavioral modules, reducing the total length even further. Over 99\% of the prompt became shared across agents regardless of industry or role, with each module carefully shortened to eliminate redundancy and self-conflicting instructions.

\subsubsection{Governed Execution}
In the final design, each module was constrained by explicit execution rules and guardrails, enabling reliable behavior while preserving generality. This architecture reflects the prompt configuration evaluated in our experiments.

A subset of these prompt variants is quantitatively evaluated in Section \ref{sec:results}, where we analyze their impact on coverage, factual accuracy, escalation behavior, and robustness.
\section{Experiments and Results}
\label{sec:results}

\begin{figure*}[!t]
    \centering
    \subfigure[\label{fig:prompt_real_estate}]{
        \includegraphics[width=0.9\textwidth]{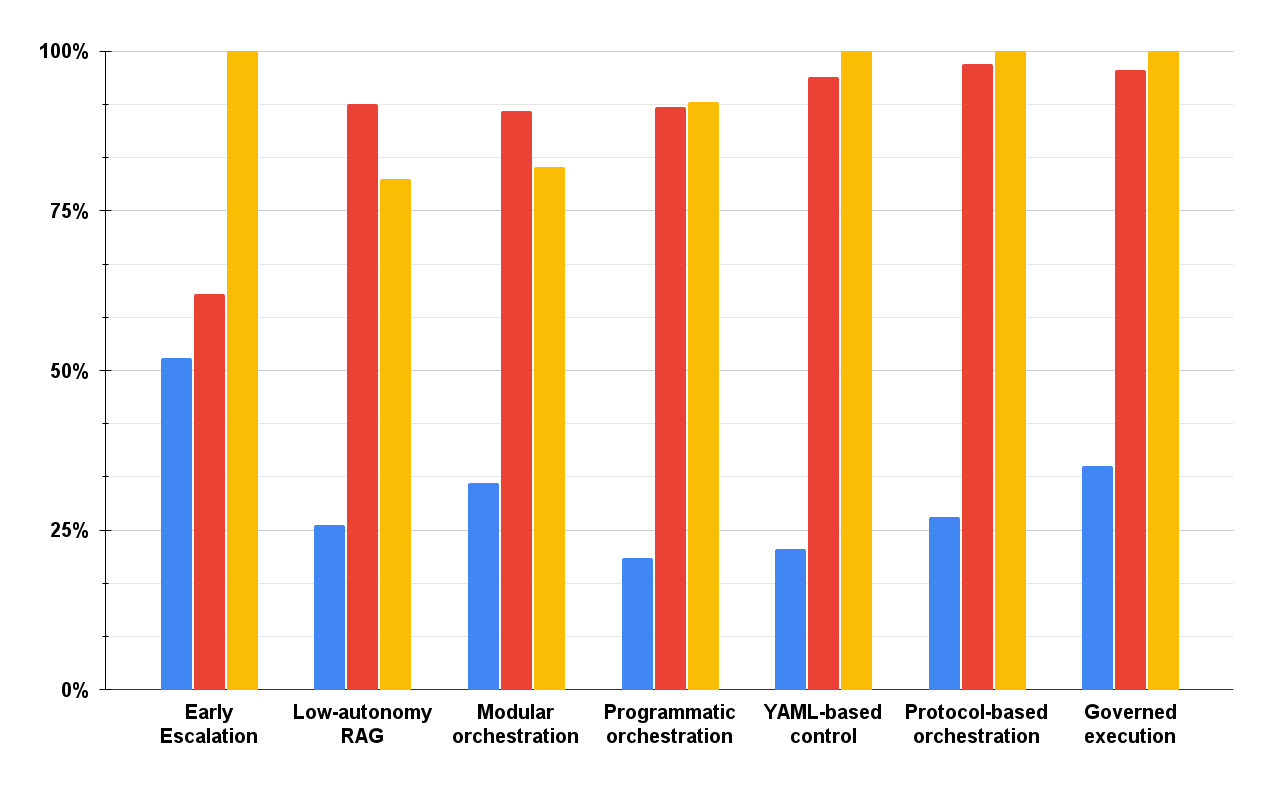}
    }
    \subfigure[\label{fig:prompt_recruitment}]{
        \includegraphics[width=0.9\textwidth]{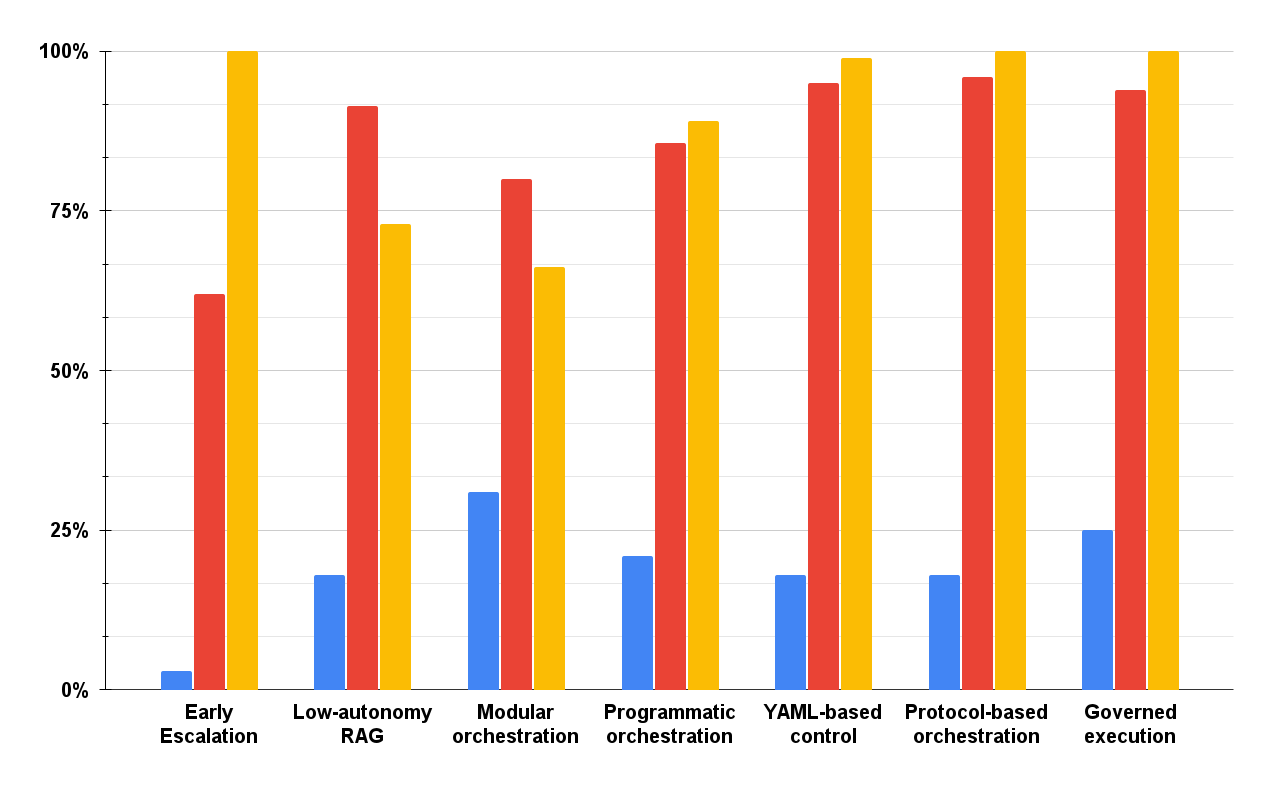}
    }
    \fbox{\includegraphics[width=0.4\textwidth]{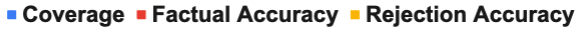}}
    \caption{Performance of different prompt orchestration strategies across two industries. Prompt variants include \emph{Early Escalation} with knowledge embedded in the system prompt, \emph{Low-autonomy RAG} with externalized knowledge, \emph{Modular orchestration} with shared step-wise actions, \emph{Programmatic orchestration} using pseudo-code actions, \emph{YAML-based} structured control, \emph{Protocol-based} orchestration with highly compact shared behavior, and \emph{Governed execution} with constrained modular actions. Coverage, factual accuracy, and rejection accuracy are reported for (a) Real Estate and (b) Specialist Recruitment.}
    \label{fig:prompt_comparison}
\end{figure*}

This section evaluates the proposed AI assistant across multiple dimensions that reflect both task performance and operational safety in realistic call center scenarios. Evaluation is conducted using a user simulator grounded in real conversational data and complemented by targeted red-teaming strategies. For each retained transcript, the underlying user intent and call objective are automatically extracted and encoded into a structured prompt that defines the simulator’s goal. The simulator then initiates interactions with the AI assistant and attempts to achieve this goal through natural language dialogue, emulating realistic user behavior observed in historical calls. By deriving simulator objectives directly from real transcripts, the evaluation preserves authentic interaction patterns and avoids reliance on synthetic or hand-crafted test cases. Results are reported separately for the Real Estate and Specialist Recruitment domains to highlight domain-specific challenges and performance characteristics.

\subsection{Coverage}

Coverage measures the proportion of calls that can be handled entirely by the AI assistant without requiring escalation to a human operator. A call is considered covered if the agent can complete the interaction by providing relevant information or appropriate guidance. This metric evaluates the practical usefulness of the agent in reducing human workload.

\subsection{Factual Accuracy}

Factual Accuracy evaluates whether the agent’s responses are correct with respect to the extracted knowledge base. Responses are compared against the ground-truth knowledge extracted from historical transcripts, assessing the effectiveness of RAG and the agent’s ability to ground its responses in verified information.

\subsection{Rejection Accuracy}

Rejection Accuracy measures the agent’s ability to recognize its limitations and correctly transfer calls to human operators when necessary. This includes scenarios where user requests are ambiguous, require real-time data, or exceed the agent’s supported scope. Correct escalation is treated as a successful outcome.

\subsection{Red-Teaming}

To evaluate the robustness and safety of the proposed AI assistant, we conduct a comprehensive red-teaming evaluation that exposes the agent to adversarial and manipulative interaction patterns. In addition to standard robustness tests, we incorporate several behavior-level attacks commonly discussed in recent LLM safety literature.

First, Out-of-Context attacks introduce irrelevant, incoherent, or nonsensical utterances that are unrelated to the ongoing conversation. These inputs test whether the agent can maintain contextual grounding and avoid hallucinated responses.

Second, Out-of-Scope attacks involve requests that exceed the agent’s supported capabilities or domain knowledge, such as tasks requiring real-time data access or unsupported services. Successful handling requires the agent to reject the request or escalate to a human operator.

Third, Prompt Injection attacks attempt to override or manipulate the agent’s internal instructions by explicitly requesting it to ignore prior rules, reveal system prompts, or adopt alternative behaviors. This includes direct requests to disclose internal prompts or to “forget” existing instructions and comply with user-defined commands. We also include Style Manipulation Attacks, in which the user requests responses in exaggerated accents, personas, or non-professional styles. While stylistic flexibility is sometimes acceptable, such requests can conflict with service constraints and professionalism requirements.

In addition, we evaluate Instruction Contradiction Attacks, where the user intentionally disagrees with or challenges information retrieved via RAG, aiming to induce inconsistency or override grounded knowledge. This tests the agent’s ability to prioritize retrieved evidence over adversarial user assertions.

Across all scenarios, the agent is expected to maintain policy compliance, preserve its predefined role, rely on retrieved knowledge when appropriate, and escalate to human operators when necessary.

\begin{table}[t]
\centering
\caption{Final Evaluation Results Across Two Industries}
\label{tab:experiment_results}
\begin{tabular}{lcc}
\toprule
\textbf{Evaluation Metrics} & \textbf{Real Estate} & \textbf{Specialist Recruitment} \\
\midrule
Coverage              & 35\% & 25\% \\
Factual Accuracy      & 97\% & 94\% \\
Rejection Accuracy    & 100\% & 100\% \\
Out-of-Context        & 98.5\% & 100\% \\
Out-of-Scope          & 98.5\% & 100\% \\
Prompt Injection      & 100\% & 100\% \\
Instruction Contradiction & 100\% & 100\% \\
\bottomrule
\end{tabular}
\end{table}

\subsection{Prompt Tuning Results}

Figure \ref{fig:prompt_comparison} illustrates how different prompt orchestration strategies influence agent performance across industries, revealing clear trade-offs between autonomy, accuracy, and operational safety. Overall, the results demonstrate that prompt structure and control granularity play a decisive role in determining the effectiveness of conversational AI agents, often outweighing domain-specific differences.

Early Escalation, which embeds detailed knowledge directly into the system prompt and prioritizes human transfer, achieves moderate coverage in Real Estate (52\%) but collapses in Specialist Recruitment (3\%). This sharp contrast highlights the fragility of monolithic, knowledge-heavy prompts when deployed in domains with higher conversational variability. While factual accuracy remains acceptable (62\% in both domains), the approach under-utilizes the agent’s potential autonomy and scales poorly across industries. The consistently perfect rejection accuracy (100\%) confirms that conservative escalation policies are effective for safety, but at the expense of coverage.

Transitioning to Low-autonomy RAG yields a substantial improvement in factual accuracy (above 91\% in both domains), validating the benefit of externalizing knowledge and grounding responses via retrieval. However, coverage remains limited (25.8\% and 18\%), and rejection accuracy drops noticeably, particularly in Specialist Recruitment (73\%). This suggests that while RAG improves correctness, excessive reliance on escalation without improved decision logic leads to premature handoffs and reduced operational efficiency.

More structured approaches, Modular orchestration and Programmatic orchestration, introduce explicit step-wise reasoning and shared action templates. These variants improve coverage stability across domains (around 30\% for Modular, ~20\% for Programmatic) while maintaining strong factual accuracy. Notably, Programmatic orchestration substantially improves rejection accuracy (above 89\%), indicating that pseudo-code–style action definitions help the agent reason more reliably about its own limitations. However, the slight drop in coverage suggests that increased structure can also constrain conversational flexibility if not carefully balanced.

The strongest overall performance emerges from YAML-based control, Protocol-based orchestration, and Governed execution. YAML-based control significantly boosts factual accuracy (up to 96\%) and restores near-perfect rejection accuracy, demonstrating the value of explicit, machine-readable structure for reducing ambiguity. Protocol-based orchestration further improves robustness, achieving consistently high factual accuracy (96–98\%) and perfect rejection accuracy across both industries, while also recovering coverage relative to earlier structured variants. These results indicate that aggressively modularizing and standardizing agent behavior—while minimizing prompt length and internal conflicts—enhances generalization across domains.

Finally, Governed execution delivers the best balance among all three metrics. It achieves the highest coverage in both industries (35\% and 25\%) while preserving high factual accuracy (97\% and 94\%) and perfect rejection accuracy. This suggests that explicitly constraining each module’s scope and authority enables the agent to act autonomously when appropriate, defer reliably when uncertain, and remain resilient to domain shifts. The consistency of results across Real Estate and Specialist Recruitment further supports the hypothesis that behavior-centric prompt design, rather than domain-specific tailoring, is the dominant factor in scalable agent deployment.

In summary, the discussion reveals a clear progression: moving from knowledge-heavy, monolithic prompts toward lean, modular, and governed orchestration consistently improves robustness, accuracy, and cross-domain transferability. These findings reinforce the view that effective conversational agents are not achieved by adding more instructions, but by carefully structuring how and when the agent is allowed to act.

\subsection{Discussion}

The experimental results provide insight into both the capabilities and limitations of the proposed conversational AI assistant when deployed in complex, real-world service domains. By evaluating performance across coverage, factual correctness, escalation behavior, and robustness to adversarial inputs, we obtain a holistic view of the system’s operational readiness.

Coverage differs notably between the two evaluated industries, with the AI assistant handling 35\% of calls in Real Estate and 25\% in Specialist Recruitment. This disparity reflects inherent domain complexity rather than system failure. Both domains rely heavily on real-time data, personalized negotiation, and context-specific constraints that are not fully captured by static knowledge bases. Specialist Recruitment exhibits lower coverage due to the frequent need for candidate-specific availability, salary negotiation, and employer-specific policies. These results highlight a key limitation of current RAG systems: while effective at delivering general knowledge, they are less suitable for tasks requiring live data access or high-stakes decision-making. Importantly, even partial coverage at these levels represents a meaningful reduction in human workload for domains traditionally considered difficult to automate.

Factual Accuracy remains high across both domains, achieving 97\% in Real Estate and 94\% in Specialist Recruitment. This demonstrates that when the agent elects to respond, it reliably grounds its answers in the extracted knowledge base. The slight degradation in Specialist Recruitment may be attributed to more nuanced role definitions and rapidly evolving terminology. These results validate the decision to separate knowledge storage from the system prompt and rely on RAG, which minimizes hallucination and supports scalable knowledge updates.

Rejection Accuracy reaches 100\% in both industries, indicating that the agent consistently recognizes its operational boundaries and escalates interactions appropriately. This is a critical safety property, as incorrect automation in these domains could lead to misinformation or service failures. The perfect rejection score suggests that the prompt design and decision logic effectively prioritize conservative behavior over overconfidence.

The Red-Teaming evaluations further demonstrate the robustness of the system. High success rates against Out-of-Context and Out-of-Scope inputs (98.5–100\%) indicate that the agent maintains conversational grounding and resists being coerced into unsupported behaviors. Perfect performance against Prompt Injection and Instruction Contradiction attacks confirms that the agent preserves system-level instructions and prioritizes retrieved evidence over adversarial user assertions. These findings suggest that the combination of structured prompts, role clarity, and RAG-based grounding provides strong defense against common LLM attack vectors.

Overall, the results suggest that the proposed framework is well-suited as a human-assistive system rather than a full replacement. By reliably handling routine inquiries, rejecting unsafe requests, and deferring complex cases, the AI assistant reduces operational burden while maintaining service quality. Future improvements in real-time data integration and domain-specific tooling are likely to further increase coverage without compromising safety.

\section{Conclusion}

This paper presented a structured framework for constructing conversational AI assistants from historical call transcripts, integrating transcript grading, strategic filtering, knowledge extraction, RAG, and systematic prompt tuning. By prioritizing transcript quality and maximizing topic coverage, the proposed approach enables the development of AI agents that respond using grounded, human-derived knowledge rather than relying on parametric memory alone. Experiments conducted in the Real Estate and Specialist Recruitment domains demonstrate that the resulting agents achieve high factual accuracy, reliable escalation behavior, and strong robustness against adversarial inputs, while safely handling a meaningful subset of real-world calls.

A central finding of this work is the critical role of prompt tuning in determining agent performance. Beyond knowledge quality, the structure, length, and internal consistency of the system prompt strongly influence coverage, accuracy, and robustness. The progression from monolithic and knowledge-heavy prompts to lean, modular, and governed prompt designs consistently improved cross-domain generalization and operational stability. These results suggest that prompt tuning should be viewed not merely as an implementation detail, but as a core design dimension for controllable and scalable conversational agents.

Importantly, the findings reinforce the view that AI assistants are most effective when positioned as human-support systems rather than full replacements. While overall coverage remains bounded in domains requiring real-time data access or complex negotiation, perfect rejection accuracy and strong red-teaming performance indicate that the agent can reliably recognize its limitations and defer to human operators when appropriate. This balance between autonomy and restraint is essential for safe deployment in production call center environments.

Future work will focus on improving knowledge consolidation to reduce extraction-induced inconsistencies, integrating real-time and transactional data sources, and extending prompt tuning methodologies to support proactive agent roles. Together, these directions point toward a scalable and trustworthy pathway for deploying AI-assisted call center systems that meaningfully augment human operators while preserving service quality and user trust.

\bibliographystyle{IEEEtran}
\bibliography{myBib}

\end{document}